\documentclass{techreport}

\usepackage[utf8]{inputenc}
\usepackage[T1]{fontenc}
\usepackage{hyperref}
\usepackage{url}
\usepackage{booktabs}
\usepackage{amsfonts}
\usepackage{nicefrac}
\usepackage{microtype}
\usepackage{xcolor}
\usepackage{graphicx}
\usepackage{enumitem}
\usepackage{overpic}
\usepackage{float}

\newcommand{\hdr}[1]{\multicolumn{1}{r}{\shortstack[r]{#1}}}

\title{Conditional Flow Matching for Single-Neuron Electrophysiology: Capturing Multimodal Responses Across Stimuli}
\date{}

\author[1,*]{Cameron Schofield}
\author[1,*,\textdagger]{Luca Ghafourpour}
\author[2]{Philip H. Wong}
\author[2,3,4]{Costas A. Anastassiou}
\author[1]{Richard E. Turner}
\affil[1]{University of Cambridge}
\affil[2]{Cedars-Sinai Medical Center}
\affil[3]{California Institute of Technology}
\affil[4]{Archimedes AI, Athena Research Center}
\affilbreak
\affil[*]{These authors contributed equally to this work}
\affilbreak
\affil[\textdagger]{Corresponding author: ldg34@cam.ac.uk}
\affilbreak

\begin{document}

\begin{abstract}
Neurons of the brain exhibit a rich repertoire of electrophysiology dynamics with the same repeated stimulus eliciting very different voltage responses from the same cell. One common approach in biophysically detailed models is to capture this variability through ensembles of deterministic parametrizations, at a cost of hundreds of thousands of CPU hours. Existing machine learning surrogates inherit the same limitation, where a stimulus is mapped to a single voltage response. We address this challenge by learning a conditional generative model for single-neuron electrophysiology, using flow matching with a velocity field conditioned on the input current. On biophysically detailed models of two human cortical interneuron types, the generated responses closely reproduce the electrophysiological feature distributions, spike-time structure, and excitability profiles, even matching the experimental recordings from the corresponding human cortical neurons. Near the firing threshold, firing and non-firing responses coexist at the same stimulus amplitude, and at high amplitudes, ensembles may split into low- and high-firing modes near depolarization block. We show that our model recovers both modes in each case, while a neural operator baseline suppresses spiking near threshold and blurs the gap between modes at depolarization block.\looseness-1
\end{abstract}

\maketitle

\section{Introduction}\label{sec:introduction}
\vspace{-2mm}
Neurons in the mammalian brain exhibit a wide range of electrophysiological responses that are area-, type-, state-, and time-specific. This is also reflected in trial-to-trial variability where the same input to the same cell typically results in very different responses in terms of spike timing, firing rate, and burstiness~\citep{mainen1995reliability,mainen1996influence,anastassiou2011ephaptic,faisal2008noise}. Capturing such diverse dynamics faithfully is essential for any single-neuron model intended to support bio-realistic circuit-level modeling~\citep{douglas1995recurrent,buzsaki2010neural,luo2021architectures}, and especially so for the human brain, where cell types differ in transcriptomic, morphological, and electrophysiological identity~\citep{berg2021human,gouwens2020integrated,lee2023signature,siletti2023transcriptomic}.

Biophysically detailed, multi-compartment PDE models grounded in cable theory provide an avenue to study these dynamics under perturbations that are experimentally infeasible in humans~\citep{koch2004biophysics,reimann2013biophysically}, with ion channel conductances tuned by evolutionary optimization to match multimodal experimental recordings~\citep{druckmann2007novel,van2016bluepyopt,nandi2022single,buchin2022multi}. Because many parametrizations perform comparably along the Pareto front~\citep{achard2006complex}, an ensemble of ``hall-of-fame'' (HoF) parametrizations is retained for each neuron and is often treated as a representation of the neuron's variability. This substitution is imperfect: each HoF model is itself deterministic, so the ensemble reflects uncertainty in the model parametrization rather than genuine trial-to-trial variability. Obtaining it also requires on the order of hundreds of thousands of CPU hours~\citep{nandi2022single,buchin2022multi}. An alternative introduces stochasticity directly into the mechanistic model~\citep{paninski2007statistical,o2007estimation,koyama2008spike}, but this requires making assumptions on the noise source and form and relies on repeated numerical integration of the underlying dynamical system.

These limitations have motivated machine learning surrogates for single-neuron electrophysiology. Neural networks, physics-informed neural networks~\cite{Raissi2019,Raissi2017Part1,Raissi2017Part2}, and neural operators~\cite{azizzadenesheli2024neural,kovachki2023neural,berner2026principled} have been applied to canonical point-neuron models such as FitzHugh-Nagumo~\cite{fitzhugh1955mathematical,nagumo1962active} and Hodgkin-Huxley~\cite{hodgkin1952quantitative}, with promising results for predicting spiking membrane potentials~\cite{su2021deep,rudi2022parameter,ferrante2022physically,kuptsov2023discovering,werneck2023replacing,pandey2024efficient,wang2026neural,centofanti2024learning}. These approaches are largely deterministic, producing a single voltage response per stimulus, and are evaluated predominantly within the regular firing regime, with performance in the highly nonlinear transition around firing threshold less clear. \texttt{NOBLE}~\cite{ghafourpour2025noble} addresses both limitations, learning a neural operator conditioned on an embedding of interpretable electrophysiological descriptors that identifies each member of a HoF ensemble. Interpolating this embedding yields responses for models beyond the finite set obtained through optimization, something the underlying biophysical models do not support in parameter space. The operator nevertheless remains deterministic given a stimulus and embedding: variability arises from sampling the embedding space rather than from learning a conditional distribution over membrane potentials.

Generative modeling provides a natural alternative in which the voltage response distribution itself becomes the target. Rather than introducing variability through an intermediate model representation, single-neuron electrophysiology can be formulated as the conditional generation of a membrane potential $V$ given an applied current $I$, directly targeting $p(V\mid I)$. Learning this distribution requires a model to represent multimodality in spike timing and spike count, preserve translation equivariance to shifts in the applied stimulus, resolve both sharp action potentials and smooth subthreshold dynamics across multiple temporal scales, and generate a coherent full voltage trace. Denoising generative models such as diffusion and flow matching are well suited to these requirements as they generate complete high-dimensional samples from noise while conditioning on auxiliary inputs~\citep{ho2020denoising,song2021score,lipman2023flow}. In particular, flow matching~\citep{lipman2023flow} learns a continuous velocity field that transports samples from a simple source distribution to the target data distribution. Conditioning this velocity field on the time-aligned applied current yields a direct model of $p(V\mid I)$, where different samples from the source noise produce different membrane potential trajectories for the same stimulus. Variability is therefore represented directly within the response distribution, without requiring an explicit neuron-model embedding. This distinction matters most where the conditioning input places the underlying deterministic dynamics near a bifurcation. There, the response distribution becomes genuinely multimodal, and a surrogate trained with a pointwise objective moves towards an intermediate response that belongs to neither mode.\looseness-1

\paragraph{Contributions}
\vspace{-3mm}
\begin{itemize}[leftmargin=*, itemsep=4pt, topsep=4pt, parsep=0pt]
    \item \textbf{We formulate single-neuron electrophysiology as conditional generative modeling.} Using flow matching with a velocity field conditioned on the time-aligned input current, we learn $p(V \mid I)$ directly. Response variability is directly targeted when fitting the surrogate, rather than being handled in an indirect and ad hoc way by an intermediate representation.

    \item \textbf{We show that our generative model captures multimodal response distributions that deterministic surrogates fail to represent.} Near firing threshold, where firing and non-firing responses coexist at the same stimulus amplitude, our model reproduces both modes. In contrast, \texttt{NOBLE} suppresses spiking near threshold and shifts the apparent firing threshold towards higher currents.

    \item \textbf{We benchmark against biophysically detailed models of two distinct human interneuron types.} Across input current amplitudes, the generated responses reproduce electrophysiological feature distributions, spike-time structure, and F-I relationships, with consistent performance across dynamical regimes.

    \item \textbf{We validate our evaluations with real human cortex recordings, rather than simulation alone.} Key statistics of the generated responses are compared directly with experimental recordings from the corresponding neurons and match qualitatively.

    \item \textbf{We generalize to unseen stimulation protocols.} Conditioning on the full input waveform, a model trained on complex time-varying stimuli generalizes to entirely new families of input stimuli.
\end{itemize}

\begin{figure}[htbp]
  \centering
  \includegraphics[width=\textwidth]{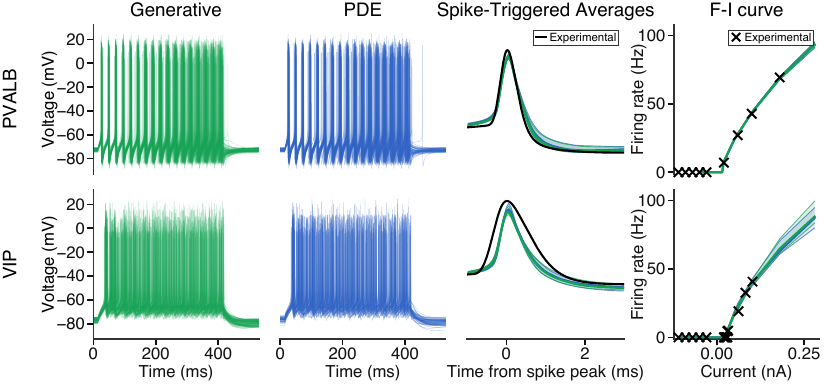}
  \caption{\textbf{Performance on different neurons.} From left to right, columns show: (1) voltage traces generated by our model; (2) voltage traces generated by PDE simulations; (3) spike-triggered voltage averages from our model, PDE simulations, and experimental recordings; and (4) F-I curves from our model, PDE simulations, and experimental recordings. For columns (3) and (4), the solid line represents the median, and the shaded area represents the interquartile range of the responses.}
  \label{fig:pvalb_vip_results_multicolumn}
\end{figure}
\section{Bio-Realistic Neuron Modeling}\label{sec:background}
\vspace{-2mm}
The biophysical single neuron PDE models are built using high resolution morphological reconstructions of human cortical neurons from Patch-seq datasets~\citep{berg2021human,gouwens2020integrated,lee2023signature}. For simulation, the models are instantiated via the BioNet interface~\citep{gratiy2018bionet} in the NEURON environment~\citep{hines2001neuron}. In each PDE model, active ion channels are placed in an ``all-active'' configuration~\citep{buchin2022multi,nandi2022single}, distributing conductances across both somatic and dendritic compartments throughout the full morphology.

The multi-objective evolutionary optimization is organized in two stages~\citep{druckmann2007novel,van2016bluepyopt,nandi2022single}. In the first stage, passive subthreshold membrane properties are optimized against non-spiking electrophysiological responses. In the second stage, active ion channel conductances are included in the optimization to account for spiking dynamics and the frequency-current (F-I) curve. In each generation of optimization, candidate HoF PDE models are ranked by mean z-score error across a biologically relevant set of electrophysiological features. After 250 generations, the best-performing parameter sets across all generations are retained as the HoF ensemble.

To evaluate the benefits of generative modeling, we focus on one parvalbumin-positive (PVALB) neuron and one vasoactive intestinal peptide (VIP) neuron from an inhibitory interneuron Patch-seq dataset of the human middle temporal gyrus~\citep{lee2023signature}. PVALB neurons are specialized fast-spiking interneurons that orchestrate high-frequency gamma oscillations (30--80\,Hz)~\citep{kim2015cortically}. VIP neurons are often found in disinhibitory circuit motifs where they target other inhibitory interneurons and release excitatory cells from dendritic inhibition~\citep{pi2013cortical,pronneke2015characterizing,veit2023cortical}. These two cell types exhibit qualitatively distinct firing regimes. While PVALB neurons coordinate high-frequency rhythms, VIP neurons operate on a slower timescale \cite{gouwens2020integrated,tremblay2016gabaergic}, allowing us to test the generalization of our framework across electrophysiological phenotypes.
\section{Method}\label{sec:method}
\vspace{-2mm}
\subsection{Problem Formulation}\label{sec:problem-formulation}
\vspace{-2mm}
We aim to learn a conditional generative model $p_\theta(V \mid I)$ that approximates the distribution of the membrane potential $V$ given an applied current $I$, where both are discretized traces in $\mathbb{R}^T$. To capture variability, we treat the responses from an ensemble of HoF models as samples from $p(V \mid I)$. We construct the dataset $\mathcal{D} = \{(I_i, V_i)\}_{i = 1}^N$ by simulating the voltage responses of randomly paired input currents and HoF models. Our model captures ensemble variability through source noise and conditions on the full current waveform to produce time-aligned voltage responses. Since the full stimulus waveform is available at inference time, we generate the voltage response jointly rather than autoregressively, allowing temporal dependencies to be represented across the complete response.

\subsection{Conditional Flow Matching}
\begin{figure}[htbp]
  \centering
  \includegraphics[width=\textwidth]{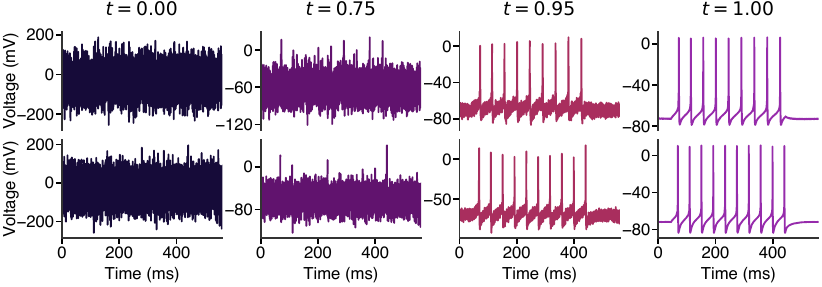}
  \caption{\textbf{Flow matching transforms noise to data.} Columns show: (1) a draw $X_0$ from the pointwise Gaussian noise source; (2) and (3) intermediate states $X_t$ at selected flow times $t$; and (4) the generated trace $X_1$.}
  \label{fig:noise_to_data_flow}
\end{figure}

\vspace{-2mm}
Let $q$ be a distribution over $\mathbb{R}^T$. A continuous normalizing flow~\cite{chen2018neural} transforms a tractable prior $p_0$ to $p_1 \approx q$ using a time-dependent velocity field $u_t$, which generates a flow $\phi_t$ through the ordinary differential equation
\begin{equation}
    \frac{d}{dt} \phi_t(x) = u_t(\phi_t(x)), \qquad \phi_0(x) = x.
\end{equation}
Writing $X_t = \phi_t(X_0)$ for $X_0 \sim p_0$, the marginals $p_t$ of this process form a probability path from $p_0$ to $p_1$ which $u_t$ generates. Samples from $p_t$ are generated by drawing $X_0 \sim p_0$ and solving the initial value problem (Figure~\ref{fig:noise_to_data_flow}).

The objective of flow matching is to fit a model $u_t^\theta$ to the velocity field that targets $p_1 = q$. This model can be trained with the Conditional Flow Matching (CFM)~\cite{lipman2023flow} objective 
\begin{equation}
    \mathcal{L}_{\mathrm{CFM}}(\theta) = \mathbb{E}_{t, X_1, X_t}
    \big\| u_t^\theta(X_t) - u_t(X_t \mid X_1) \big\|^2 ,
    \qquad X_1 \sim q, \;\; X_t \sim p_t(\cdot \mid X_1) ,
\end{equation}
which constructs $p_t$ as a mixture of conditional paths $p_t(\cdot \mid X_1)$ with conditional velocities $u_t(\cdot \mid X_1)$.

We use the linear conditional path~\cite{lipman2023flow,albergo2023stochastic}, taking $p_0 = \mathcal{N}(\mathbf{0}, \mathbf{I}_T)$ and $X_t = (1 - t) X_0 + t X_1$. This gives the target velocity
\begin{equation}
    u_t(X_t \mid X_0, X_1) = X_1 - X_0,
\end{equation}
which is constant in flow time $t$.

The time-aligned conditioner is added as an additional input to the velocity network $u_t^\theta(X_t \mid I)$. Taking $X_1 = V$ and training pairs $(I, V)$, the minimizer of the CFM objective is the marginal velocity field generating $p(V \mid I)$ \cite{lipman2024guide}.

\subsection{Training Details}
\vspace{-2mm}
We parameterize the velocity field $u_t^\theta$ with a five-resolution one-dimensional U-Net~\cite{ronneberger2015unet}, using dilated convolutions at the bottleneck to capture long-range temporal dependencies. The input current is encoded and injected at each resolution, while flow time is embedded sinusoidally. Models are trained with AdamW~\cite{loshchilov2019decoupled} for 200 epochs and sampled using 20 midpoint steps. Full architectural, optimization, and sampling details, together with associated ablations, are provided in Appendix~\ref{app:architecture}.

As reported in~\cite{ghafourpour2025noble}, the numerical solver takes 36,200 seconds to generate 10,000 voltage responses.
We record the time necessary to generate 10,000 samples on a single NVIDIA A100 GPU using our trained generative model. Each sample requires 40 network evaluations. When doing one inference at a time, we generate 10,000 samples in 6,600 seconds, a speedup of approximately $5.5\times$. Using a batch size of 1000, the model generates 10,000 samples in approximately 865 seconds, a speedup of approximately $42 \times$.  Inference cost can be reduced further, either by taking fewer network evaluations per
sample or by distilling the learned velocity field into a few-step generator~\cite{song2023consistency,geng2026mean,salimans2022progressive}, in both cases at a fixed cost in fidelity.

\section{Results}\label{sec:results}
\vspace{-2mm}
In the following subsections, we compare our trained generative model against the HoF models, \texttt{NOBLE} as trained by the original authors and released at~\url{https://github.com/neuraloperator/noble/}, and real experimental human cortex recordings.

\vspace{-2mm}
\subsection{Recovering Response Distributions at Unseen Currents}
\vspace{-2mm}
We first assess the generated ensemble of full voltage traces at unseen current amplitudes. Figure~\ref{fig:pvalb_vip_results_multicolumn} qualitatively showcases that for both the PVALB and VIP neurons, the generated ensembles closely resemble those produced by the HoF models, capturing both the response dynamics and their variability. To compare spike shape more directly, we compute spike-triggered averages, where the median and interquartile range of the generated responses closely match those of the HoFs, while the experimental curves lie within the same range. We observe similar agreement in the F-I curves, where the generated firing-rate distributions track the HoFs across stimulation amplitudes and the experimental measurements align closely with the median response. These results indicate that our model recovers the underlying conditional response distribution at unseen currents.

\begin{figure}[htbp] 
    \centering 
    \begin{minipage}[c]{0.65\textwidth} 
        \centering 
        \begin{overpic}[width=\linewidth]{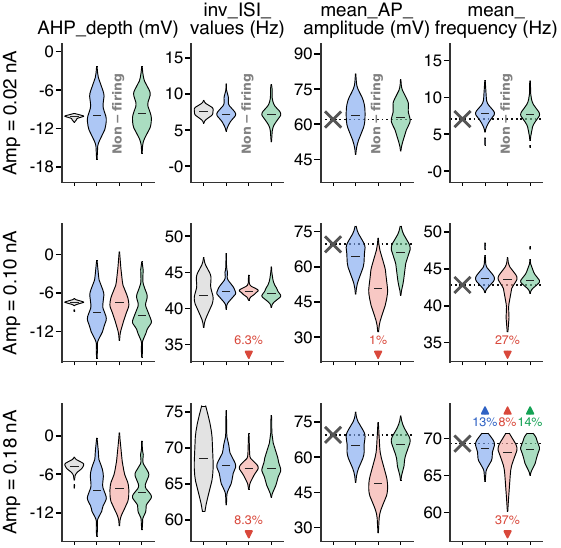}
            \put(0, 95){\textsf{\textbf{\LARGE A}}}
        \end{overpic}
    \end{minipage}%
    \hfill
    \begin{minipage}[c]{0.33\textwidth} 
        \centering 
        \begin{overpic}[width=\linewidth]{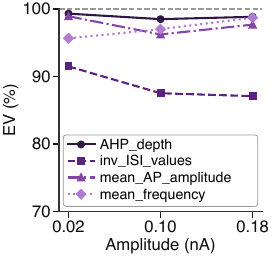}
            \put(-5, 89){\textsf{\textbf{\LARGE B}}}
        \end{overpic}
        
        \vspace{0.1cm} 
         
        \begin{overpic}[width=\linewidth]{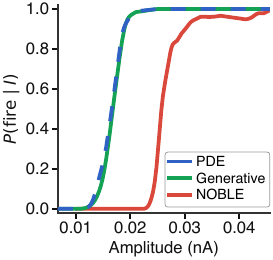}
            \put(-5, 89){\textsf{\textbf{\LARGE C}}}
        \end{overpic}
    \end{minipage} 
     
    \caption{\textbf{PVALB feature distributions and firing probability.} \textbf{A)} Distribution of key electrophysiological features for a PVALB neuron from experimental recordings (grey), PDE simulations (blue), \texttt{NOBLE} (red), and our model (green) across three representative amplitudes. Arrows indicate the fraction of values outside of the displayed axis range. \textbf{B)} Explained variance (EV, \%) across quantile levels of generated and PDE distributions. \textbf{C)} $P(\mathrm{fire} \mid I)$, the empirical firing probability given current amplitude for PDE simulations (blue; dashed), \texttt{NOBLE} (red), and our model (green).} 
    \label{fig:pvalb_combined} 
\end{figure}

To assess whether the model preserves electrophysiological structure beyond individual voltage traces, we compute a set of electrophysiological features for each response (see Appendix~\ref{app:ephys_definitions} for definitions). These features are robust to small temporal shifts in voltage traces that can otherwise produce large pointwise errors, providing a more representative measure of electrophysiological similarity. Figure~\ref{fig:pvalb_combined}A qualitatively compares the resulting feature distributions with \texttt{NOBLE}, the HoF models, and experimental recordings. Our generated distributions closely match the HoFs across features and stimulation amplitudes. A rather clear distinction from \texttt{NOBLE} appears in the firing threshold regime. At 0.02nA, the HoF data contain both firing and non-firing responses, which our model captures, while \texttt{NOBLE} produces no firing responses at all. We add to Appendix~\ref{app:more_pvalb_results} the raw ensemble of traces in Figure~\ref{fig:pvalb_representative_traces}, which showcases this failure to capture the bimodality present in the firing regime.\looseness-1

We quantify the agreement between the feature distributions generated by the HoF models and our generative model in Figure~\ref{fig:pvalb_combined}B using explained variance. Our model captures most of the variability in the HoF feature distributions, with explained variance above 95\% for several features across amplitudes. Performance is lower for inverse inter-spike interval (ISI) values, consistent with Figure~\ref{fig:pvalb_combined}A. Figure~\ref{fig:pvalb_distributional_alignment} separates this agreement into location and spread. The normalized mean bias stays within about $0.25\sigma$ for every feature and amplitude. The largest discrepancy is in dispersion, with generated inter-spike intervals under-dispersed at 0.10nA, consistent with the lower explained variance for inverse ISI.

Figure~\ref{fig:pvalb_combined}C highlights the current shortcoming of \texttt{NOBLE} in regimes where both firing and non-firing responses coexist by plotting firing probability as a function of stimulation amplitude. Our model closely follows the HoF distribution, whereas \texttt{NOBLE} shows a clear shift towards higher firing thresholds. One possible explanation is its deterministic relative $\mathrm{L4}$ objective. Near threshold, the HoF distribution contains both firing and non-firing responses at the same stimulation amplitude. \texttt{NOBLE} minimizes a pointwise relative $\mathrm{L4}$ objective over this multimodal distribution, which can prefer an intermediate response rather than either mode. Ultimately, this can suppress spiking near threshold and shift the predicted firing threshold towards higher currents. The firing threshold is precisely the regime where the underlying dynamics sit near a bifurcation, which is why it discriminates between the two approaches more sharply than either the sub- or supra-threshold regime.\looseness-1

Table~\ref{tab:pvalb_performance_by_dynamical_regime} quantifies the agreement between the generated and HoF feature distributions using the Wasserstein-2 distance. Overall, the distances remain small across features relative to the underlying variance in feature distributions. However, since averaging across all amplitudes could hide regime-specific failure modes, we repeat the analysis separately across three firing regimes: threshold (where firing and non-firing responses coexist across all HoFs), low-firing (with mean spike counts between 5 and 17) and high-firing (with mean spike counts between 18 and 37). Performance remains consistent across these regimes, with no regime-specific failure mode. The threshold regime is noticeably more challenging for several features, but the overall agreement remains strong.\looseness-1

Corresponding results for the VIP neuron are provided in Appendix~\ref{app:vip_results}.\looseness-1

\begin{table}
\centering
\begin{tabular}{lrrrrr}
\toprule
Regime
    & \hdr{AP\\width\\($\mu$s)}
    & \hdr{AHP\\depth\\(mV)}
    & \hdr{mean AP \\amplitude\\(mV)}
    & \hdr{inv ISI\\(Hz)}
    & \hdr{mean \\frequency\\(Hz)}
    \\
\midrule
All & 25.3 & 0.595 & 1.14 & 0.533 & 0.334 \\
Threshold & 43.7 & 1.186 & 2.23 & 0.509 & 0.783 \\
Low-firing & 27.0 & 0.533 & 1.22 & 0.419 & 0.240 \\
High-firing & 22.9 & 0.592 & 1.03 & 0.605 & 0.361 \\
\bottomrule
\end{tabular}
\caption{\textbf{Performance by dynamical regime.} Wasserstein-2 distances between reference and generated feature distributions, averaged across amplitudes in different regime types. These regimes are: (1) threshold amplitudes, where at least one HoF model produces a steady state response and at least one HoF model produces a firing response, (2) low-firing amplitudes, which produce mean spike counts between 5 and 17, and (3) high-firing amplitudes, which produce mean spike counts between 18 and 37.}
\label{tab:pvalb_performance_by_dynamical_regime}
\end{table}

\vspace{-1mm}
\subsection{Capturing Spike-Timing Variability}
\vspace{-2mm}
\begin{figure}[htbp]
    \centering
    \includegraphics[width=\textwidth]{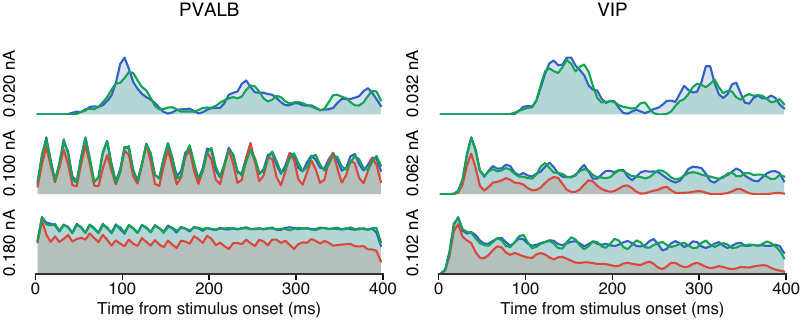}
    \caption{\textbf{Spike-timing variability.} Spike peak timing at representative PVALB and VIP current amplitudes, comparing smoothed spike-time densities from PDE simulations (blue), Generative (green), and \texttt{NOBLE} (red) voltage responses.}
    \label{fig:pvalb_spike_peak_timing}
\end{figure}
Matching electrophysiological feature distributions alone does not ensure that the temporal structure of spiking is preserved. Spike timing determines when information is transmitted to downstream neurons, and can thus affect the resulting network dynamics even when summary electrophysiological features such as firing rate or spike width match. In Figure~\ref{fig:pvalb_spike_peak_timing} we show the distribution of spike times across representative current amplitudes for the PVALB and VIP neuron. Our generative model closely follows the HoF spike-time densities across regimes, while \texttt{NOBLE} shows larger discrepancies, and as seen before, does not capture spikes at amplitudes within the threshold regime.\looseness-1

\vspace{-1mm}
\subsection{Capturing Multimodal Dynamics}
\begin{figure}[htbp]
    \centering
    \includegraphics{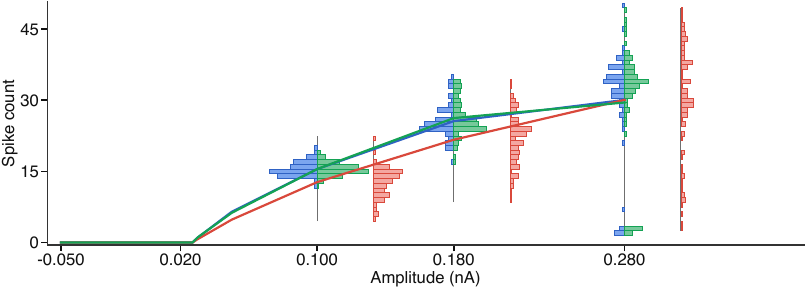}
    \caption{\textbf{Multimodal spike counts near depolarization block.} Spike count versus input current amplitude for VIP neuron. Lines plot the mean spike count in the conditional voltage responses. Histograms plot the spike count distribution for the PDE (blue), Generative (green), and \texttt{NOBLE} (red) responses at current amplitudes 0.1nA, 0.18nA, and 0.28nA.}
    \label{fig:vip_spikecount_curve_with_overlays}
\end{figure}

\vspace{-2mm}
At sufficiently high current amplitudes, the VIP HoF ensemble approaches depolarization block. Differences across HoF parametrizations place models on either side of this transition, with some continuing to fire while others enter a low-firing state. This produces a multimodal conditional response distribution. Figure~\ref{fig:vip_spikecount_curve_with_overlays} shows this clearly at 0.28nA, where the HoF spike-count distribution separates into distinct low- and high-firing modes. Our generative model recovers both modes and their approximate probability mass, whereas \texttt{NOBLE} spreads mass across the range and resolves neither mode. Importantly, this difference would be obscured by the mean response alone. As such, our model can qualitatively represent distinct dynamical modes under the same conditioning input rather than collapsing them into a single response.

\subsection{Generalization to Time-Varying Stimulation Protocols}
\vspace{-2mm}
While the previous sections consider DC-step inputs, conditioning on the full time-aligned stimulus waveform allows the model to handle more general stimulation protocols. We therefore generated a new dataset of time-varying current waveforms (Appendix~\ref{app:data}) and trained a separate model on this broader stimulus distribution, using the same model, optimizer, and training hyperparameters as the model trained on DC step waveforms. 

Table~\ref{tab:pvalb_augmented_performance_by_dynamical_regime} repeats the dynamical-regime evaluation of Table~\ref{tab:pvalb_performance_by_dynamical_regime}, but using the model trained on more complex time-varying stimuli. Performance remains comparable across threshold, low-firing, and high-firing regimes, with no new regime-specific failure mode. Hence, expanding the training distribution to time-varying inputs does not substantially degrade the model's ability to reproduce the response distributions elicited by DC steps.

\begin{table}[!h]
\centering
\begin{tabular}{lrrrrr}
\toprule
Regime
    & \hdr{AP\\width\\($\mu$s)}
    & \hdr{AHP\\depth\\(mV)}
    & \hdr{mean AP\\amplitude\\(mV)}
    & \hdr{inv ISI\\(Hz)}
    & \hdr{mean\\frequency\\(Hz)}
    \\
\midrule
All & 27.9 & 0.786 & 1.60 & 0.627 & 0.378 \\
Threshold & 43.6 & 1.120 & 2.87 & 0.767 & 0.962 \\
Low-firing & 26.8 & 0.865 & 1.59 & 0.620 & 0.289 \\
High-firing$^\dagger$ & 27.1 & 0.658 & 1.45 & 0.623 & 0.404 \\
\bottomrule
\end{tabular}
\caption{\textbf{Performance by dynamical regime when trained on time-varying stimuli.} \\ Wasserstein-2 distances between reference and generated feature distributions, averaged across square pulse currents in different regime types. $^\dagger$High-firing amplitude range is cut off at 0.18nA to match the max amplitude in the time-varying stimuli dataset.}
\label{tab:pvalb_augmented_performance_by_dynamical_regime}
\end{table}

\begin{figure}[htbp]
    \centering
    \includegraphics[width=\textwidth]{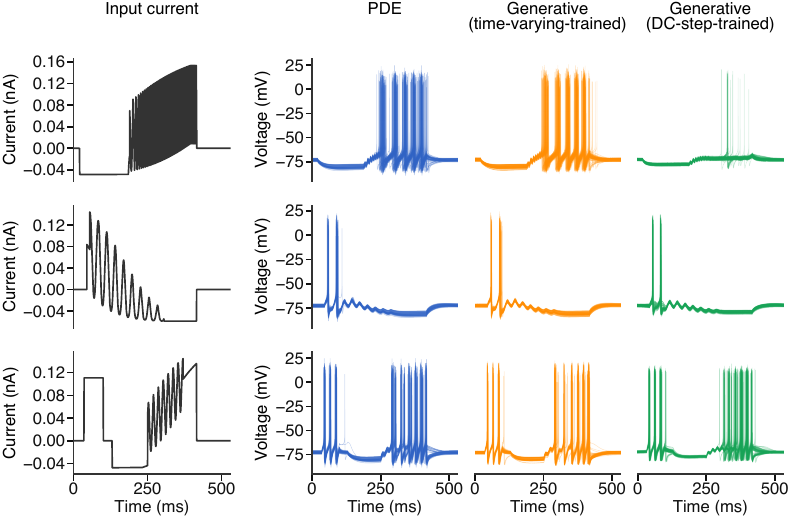}
   \caption{\textbf{Generalization to unseen time-varying stimuli.} Predicted PVALB responses to unseen time-varying stimuli, comparing PDE simulations with generative models trained on time-varying and DC-step currents.}
    \label{fig:pvalb_selected_time_varying_protocols}
\end{figure}

We next evaluate both models on unseen time-varying stimuli. Figure~\ref{fig:pvalb_selected_time_varying_protocols} compares their generated response ensembles on representative held-out waveforms. The DC-trained model generalizes to some non-DC stimuli by producing spikes in the correct regions, but blur in the generated ensembles indicates poor ISI consistency. Because the same kernels apply everywhere, any input that locally resembles a DC step within the kernel's receptive field is indistinguishable from a step, so the model stitches locally step-like predictions across the trace. This provides good local behavior, but poor global consistency. Figure~\ref{fig:pvalb_frequency_sweep} demonstrates that the performance of the DC-trained model on the distribution of ISIs suffers more when the input frequency exceeds 40 Hz, at which point the kernel weights learned for locally step-like behavior are no longer suitable. This failure motivates training on time-varying stimuli. Figure~\ref{fig:pvalb_selected_time_varying_protocols} and Figure~\ref{fig:pvalb_more_time_varying} qualitatively show that our model, when trained on complex time-varying currents, can capture the time-varying voltage dynamics of the PDE simulations. This requires no architectural change and allows the model to generalize across stimulus families.\looseness-1

\newpage
\section{Conclusion}\label{sec:conclusion}

We introduced a conditional flow matching model of voltage dynamics that learns $p(V \mid I)$ directly, representing variability directly without an ad hoc distribution over latent feature embeddings. Trained on ensembles of PDE simulations for PVALB and VIP neurons, it recovers the response distribution at unseen currents, captures temporal dependencies and multimodal dynamics, and transfers across stimulus families. These properties make conditional generative surrogates a promising route toward large-scale circuit modeling, where biophysically detailed simulations become prohibitively expensive. Extending the framework from single neurons to interacting populations, while using distillation or few-step generation to approach the speed of deterministic surrogates such as \texttt{NOBLE}, are natural directions for future work.

\section*{Acknowledgments}
C.A.A. is supported by the National Institutes of Health R01-NS120300 and R01-NS130126. P.H.W. is supported by the National Institutes of Health R01-NS130126.

\bibliographystyle{unsrtnat}
\bibliography{References}
\clearpage
\appendix
\section{Electrophysiological Features}\label{app:ephys_definitions}
For electrophysiological feature extraction and metrics, we use code defined in the Electrophys Feature Extraction Library (eFEL) available at \url{https://github.com/BlueBrain/eFEL}. The formulas, codes, and definitions about each electrophysiological feature can be found at \url{https://efel.readthedocs.io/en/latest/eFeatures.html}.

Below, we list the electrophysiological features considered in this work (where AP denotes action potential and AHP denotes after-hyperpolarization). We use the default spike-detection threshold of -20 mV and pool vector-valued features when comparing distributions.

{\setlength{\leftmargini}{2em}
 \setlength{\itemsep}{1em} 
\begin{itemize}
    \item \texttt{AHP\_depth}: Collection of relative voltage values after each AHP

    \item \texttt{AP\_width}: Collection of widths of each spike computed at the spike-detection threshold

    \item \texttt{inv\_ISI\_values}: 1.0 over all interspike intervals
    
    \item \texttt{mean\_AP\_amplitude}: The mean of all of the AP amplitudes

    \item \texttt{mean\_frequency}: The mean frequency of the firing rate

    \item \texttt{spikecount}: Number of spikes in the trace, including outside of stimulus interval

\end{itemize}}

\section{Model Architecture and Training Details}\label{app:architecture}

We evaluated design choices on a small dataset of voltage responses for a single PVALB neuron, obtained by producing all 60 HoF responses at 119 fixed current amplitudes spanning the range $[-0.11, 0.28]$nA.

\subsection{Velocity Architecture}

We compared the U-Net against DiffWave~\cite{kong2021diffwave}, a non-autoregressive adaptation of WaveNet that operates at full temporal resolution using a deep stack of dilated convolutions to grow the receptive field. The U-Net was selected because it was far better at preserving global consistency in generated traces; the DiffWave model struggled to generate consistent ISIs (Table~\ref{tab:arch_w2}) because of a weaker global representation. At approximately matched parameter counts (4,973,569 for U-Net and 5,225,569 for DiffWave), the U-Net was additionally three times faster per network evaluation.

\begin{table}[h]
\centering
\begin{tabular}{@{}lrrrrr@{}}
\toprule
    & \hdr{AP\\width\\($\mu$s)}
    & \hdr{AHP\\depth\\(mV)}
    & \hdr{mean AP \\amplitude\\(mV)}
    & \hdr{inv ISI\\(Hz)}
    & \multicolumn{1}{r@{}}{\shortstack[r]{Spike\\count}} \\
\midrule
DiffWave & \textbf{30.1} & 0.99 & \textbf{3.08} & 3.53 & 0.48 \\
U-Net    & 49.3 & \textbf{0.87} & 4.19 & \textbf{0.76} & \textbf{0.25} \\
\bottomrule
\end{tabular}
\caption{Wasserstein-2 distances between reference and generated feature distributions, averaged over amplitudes, for the DiffWave and U-Net models.}
\label{tab:arch_w2}
\end{table}

\begin{figure}[htbp]
    \centering
    \includegraphics{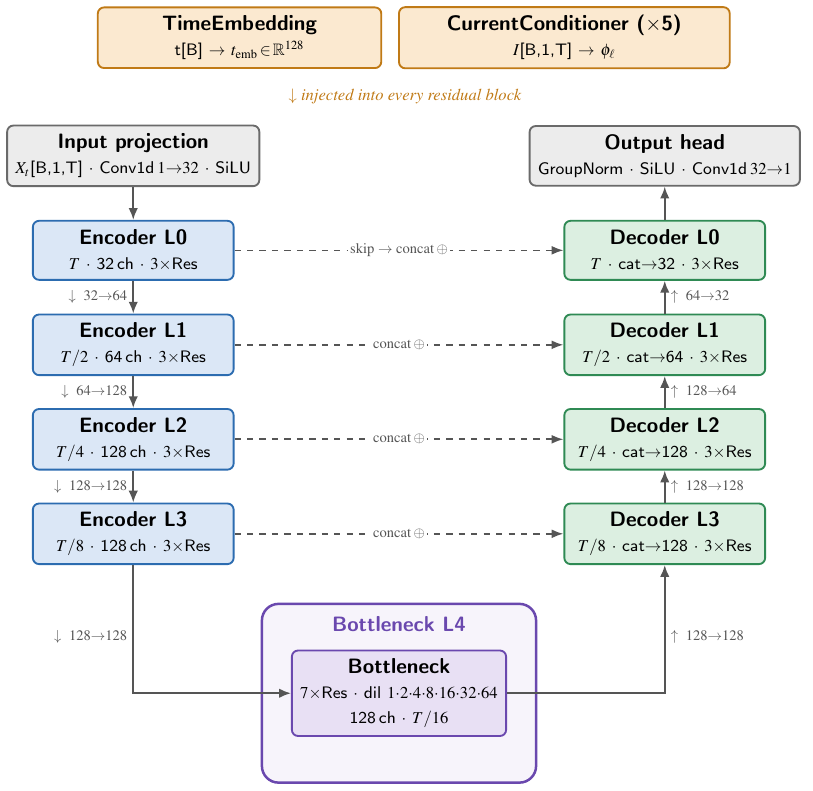}
    \caption{U-Net model for predicting the flow matching velocity from the interpolated voltage state $X_t$, flow time $t$, and input current $I$. Resolution is halved at each encoder stage and restored by interpolation in the decoder; skip connections carry full-resolution features around the bottleneck. Flow time and current conditioning enter every residual block.}
    \label{fig:unet_architecture}
\end{figure}

The selected velocity model is a one-dimensional U-Net~\cite{ronneberger2015unet} with residual blocks at five resolutions, stride-two downsampling, and skip concatenations to a mirrored decoder (Figure~\ref{fig:unet_architecture}). At the lowest resolution, we use a stack of bidirectional dilated convolutions with geometrically increasing dilation instead of self-attention, providing a receptive field covering the full trace at linear computational cost~\cite{kong2021diffwave,oord2016wavenet}. We use replicate padding to avoid introducing a constant boundary from which convolutional filters could infer absolute position~\cite{islam2020how,kayhan2020on}. The input current is encoded by a convolutional stack, resampled to each resolution, and injected additively into every residual block. Flow time is sinusoidally embedded and likewise added to each block.

The same residual block is used at each resolution and in the bottleneck. Each block applies GroupNorm, SiLU, a convolution, and GroupNorm; adds the flow-time embedding broadcast across the trace as well as the resolution's current features; and then applies SiLU, dropout, and a second convolution. The output is the sum of this result and the input, divided by $\sqrt{2}$. 

\subsection{Training Objective}

The $\mathrm{L2}$ loss is minimized by the conditional mean that recovers the marginal velocity field required to transport probability mass from $p_0$ to $q$. The $\mathrm{L1}$ loss is minimized by the conditional median; for multimodal distributions, the minimizer is pulled toward whichever mode carries more mass, rather than averaging across divergent paths. This yields a biased field that may fail to transport the distribution to $q$.

Despite this theoretical limitation, we tested both loss functions to see if mode selection would sharpen individual traces that averaging would blur under spike-timing jitter. Empirically, $\mathrm{L1}$ reduces subthreshold noise and improves spike amplitude accuracy (Table~\ref{tab:loss_w2}). However, $\mathrm{L2}$ better preserves the firing distribution while $\mathrm{L1}$ has a greater tendency to have mode collapse. We used the $\mathrm{L2}$ loss to train our models.

\begin{table}[h]
\centering
\begin{tabular}{@{}lrrrrrr@{}}
\toprule
& \multicolumn{5}{c}{Wasserstein-2 distance, spiking amplitudes} & Subthreshold noise \\
\cmidrule(lr){2-6}\cmidrule(l){7-7}
    & \hdr{AP\\width\\($\mu$s)}
    & \hdr{AHP\\depth\\(mV)}
    & \hdr{mean AP \\amplitude\\(mV)}
    & \hdr{inv ISI\\(Hz)}
    & \hdr{Spike\\count}
    & \multicolumn{1}{r@{}}{\shortstack[r]{residual\\RMS\\(mV)}} \\
\midrule
$\mathrm{L1}$ & 62.6 & 1.33 & \textbf{2.20} & 0.80 & 0.34 & \textbf{0.07} \\
$\mathrm{L2}$ & \textbf{49.3} & \textbf{0.87} & 4.19 & \textbf{0.76} & \textbf{0.25} & 0.21 \\
\bottomrule
\end{tabular}
\caption{Wasserstein-2 distances between reference and generated feature distributions, averaged over amplitudes, for U-Net models trained with $\mathrm{L1}$ and $\mathrm{L2}$ losses. The final column is the RMS of the high-frequency residual in non-spiking traces averaged over subthreshold amplitudes, for which the reference value is $0.003$ mV (closer to this is better).}
\label{tab:loss_w2}
\end{table}

\subsection{Sampling Method}

\begin{figure}[htbp]
    \centering
    \includegraphics{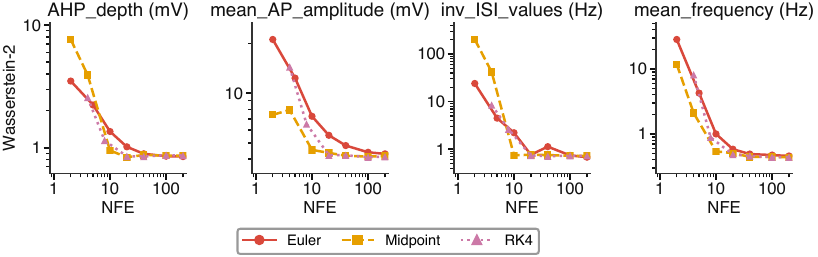}
    \caption{Wasserstein-2 distance between reference and generated feature distributions, as a function of solver and NFE. Both axes are log-scaled.}
    \label{fig:pvalb_sampling_method_nfe_w2_by_feature}
\end{figure}

We compared Euler, midpoint, and fourth-order Runge--Kutta integration at matched numbers of function evaluations (NFE) in terms of sample quality (Figure~\ref{fig:pvalb_sampling_method_nfe_w2_by_feature}). All three converge to a common limit by 200 NFE, set by error in the learned velocity field rather than in the integration. Midpoint converges fastest, reaching the common limits in 10 to 20 NFE. We sample with 20 midpoint steps (40 NFE) so that the reported results are not confounded by integration error, although fewer steps can still produce high-quality samples.

\subsection{Hyperparameters}

We performed a hyperparameter sweep on the U-Net to determine an optimal configuration. We use a base width of 32 channels, channel multipliers $(1, 2, 4, 4, 4)$, three residual blocks per resolution level, kernel size 5, group normalization with 8 groups, and dropout 0.03. The bottleneck uses dilations $(1, 2, 4, 8, 16, 32, 64)$. Flow-time embeddings use dimension 64 with hidden dimension 128.

Models are optimized using AdamW~\cite{loshchilov2019decoupled} with a base learning rate of $5\times10^{-4}$, weight decay $10^{-4}$ applied to parameters of dimension two or higher, and momentum coefficients $\beta_1=0.9$ and $\beta_2=0.999$. We use a five-epoch linear warmup with initial factor $0.2$, followed by cosine annealing~\cite{loshchilov2017sgdr} to a minimum learning rate of $10^{-5}$. An exponential moving average with rate $0.9995$ is maintained throughout training, and gradient norms are clipped at $1.0$. Models are trained for 200 epochs with batch size 32, with the checkpoint of lowest validation loss used for evaluation. All training is performed on NVIDIA A100 GPUs.

\section{Dataset Generation}
\label{app:data}

The voltage response datasets for the PVALB (with DC step inputs and time-varying inputs) and VIP (with DC step inputs) neurons are generated from a single neuron morphology and a set of 60 HoF conductance parametrizations obtained by multi-objective evolutionary optimization, of which 50 are used for training. For each model we draw 1{,}680 stimuli, giving 100{,}800 current--voltage pairs; every stimulus is paired with a single model, so a conditioner is almost never observed more than once. Responses are simulated in NEURON~\cite{hines2001neuron} at a $0.02$ms timestep over a 400 ms activation window and cropped to 45 ms before onset, the activation window, and 115 ms after it. Traces are then low-pass filtered and decimated by a factor of three, which preserves the fidelity of the extracted features while reducing computational expense~\cite{ghafourpour2025noble}, giving 9{,}334 samples at $\Delta t = 0.06$ms. Current and voltage traces are min-max scaled to $[-1, 1]$, fit on the training split. The 50 training models contribute 84{,}000 traces, split 90/10 into 75{,}600 training and 8{,}400 validation examples.

\subsection{Current Amplitude Sampling}

\begin{figure}[htbp]
    \centering
    \includegraphics{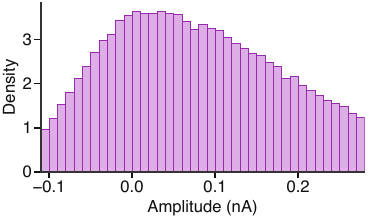}
    \caption{Empirical distribution of DC step amplitudes in $[-0.11, 0.28]$nA considered.}
    \label{fig:dc_amplitude_distribution}
\end{figure}

We use square DC step currents, the standard protocol for characterizing neuron behavior in electrophysiological experiments. Amplitudes are drawn from a truncated skew-normal distribution on $[-0.11, 0.28]$nA (Figure~\ref{fig:dc_amplitude_distribution}), the range over which the HoF models are experimentally validated. Its mode is placed within the peri-threshold window, where responses transition abruptly between non-spiking and spiking and the map from current to response is most
nonlinear, and its positive skew provides coverage of the sustained spiking regime. This distribution is the same used for the \texttt{NOBLE} training data~\citep{ghafourpour2025noble}, making the comparisons fair.

\subsection{Time-Varying Stimulus Construction}

Time-varying stimuli are constructed by summing a random number of sine, step, ramp and chirp basis components, each with independently sampled amplitude, onset and duration, and with frequency parameters additionally sampled for the oscillatory components (see Table~\ref{tab:custom_waveform_basis_distributions}). At least one ramp is always included, so no pure square pulse occurs in this dataset and evaluation on step currents is a test of transfer. The summed waveform is smoothly compressed to lie within amplitude bounds of $[-0.06, 0.18]$nA.

\begin{table}[!h]
\begin{tabular}{llll}
\toprule
Basis & Quantity & Symbol & Distribution \\
\midrule
\multicolumn{4}{@{}p{0.96\linewidth}@{}}{\textbf{Sine:} $a\sin\!\left(2\pi f(t-s)/1000 + \phi\right)$ for $s \le t \le s+d$; zero otherwise.}\\
 & count & $N$ & \(\mathrm{Uniform}\{0,1,2\}\) \\
 & amplitude & $a$ & $\mathrm{Uniform}(0.5, 1.5)$ \\
 & frequency & $f$ & $\mathrm{Uniform}(20, 80)$ Hz \\
 & phase & $\phi$ & $\mathrm{Uniform}(0, 2\pi)$ rad \\
 & start time & $s$ & $\mathrm{Uniform}(1, 200)$ ms \\
 & duration & $d$ & $\mathrm{Uniform}(50, 300)$ ms \\
\addlinespace
\multicolumn{4}{@{}p{0.96\linewidth}@{}}{\textbf{Step:} constant value $a$ for $s \le t \le s+d$; zero otherwise.}\\
 & count & $N$ & \(\mathrm{Uniform}\{0,1,2\}\) \\
 & amplitude & $a$ & $\mathrm{Uniform}(-1, 1)$ \\
 & start time & $s$ & $\mathrm{Uniform}(1, 100)$ ms \\
 & duration & $d$ & $\mathrm{Uniform}(50, 300)$ ms \\
\addlinespace
\multicolumn{4}{@{}p{0.96\linewidth}@{}}{\textbf{Ramp:} $m(t-s)$ for $s \le t \le s+d$, then held at $md$ for the remainder of the activation window.}\\
 & count & $N$ & \(\mathrm{Uniform}\{1,2\}\) \\
 & slope & $m$ & $\mathrm{Uniform}(-0.015, 0.015)$ generator units/ms \\
 & start time & $s$ & $\mathrm{Uniform}(1, 100)$ ms \\
 & duration & $d$ & $\mathrm{Uniform}(30, 300)$ ms \\
\addlinespace
\multicolumn{4}{@{}p{0.96\linewidth}@{}}{\textbf{Chirp:} $a\sin\!\left(2\pi[f_0\tau + r\tau^2/2]\right)$ for $t \ge s$, where $\tau=(t-s)/1000$ s.}\\
 & count & $N$ & \(\mathrm{Uniform}\{0,1\}\) \\
 & amplitude & $a$ & $\mathrm{Uniform}(0.5, 1.0)$ \\
 & start time & $s$ & $\mathrm{Uniform}(1, 100)$ ms \\
 & frequency ramp rate & $r$ & $\mathrm{Uniform}(1000, 2000)$ Hz/s \\
 & base frequency & $f_0$ & $\mathrm{Uniform}(50, 100)$ Hz \\
\bottomrule
\end{tabular}

\caption{Construction of the custom current waveforms. For each waveform, the number of components of each basis type is sampled first from the discrete distribution shown. Conditional on a component being included, its parameters are sampled independently, each from the listed distributions. Times are relative to the activation window onset.}
\label{tab:custom_waveform_basis_distributions}
\end{table}

\clearpage

\begin{figure}[htbp]
\section{Supplementary Results for the PVALB Neuron}\label{app:more_pvalb_results}

    \subsection{Voltage Trace Ensembles}
    \vspace{0.2cm}
    \centering
    \includegraphics[width=\textwidth]{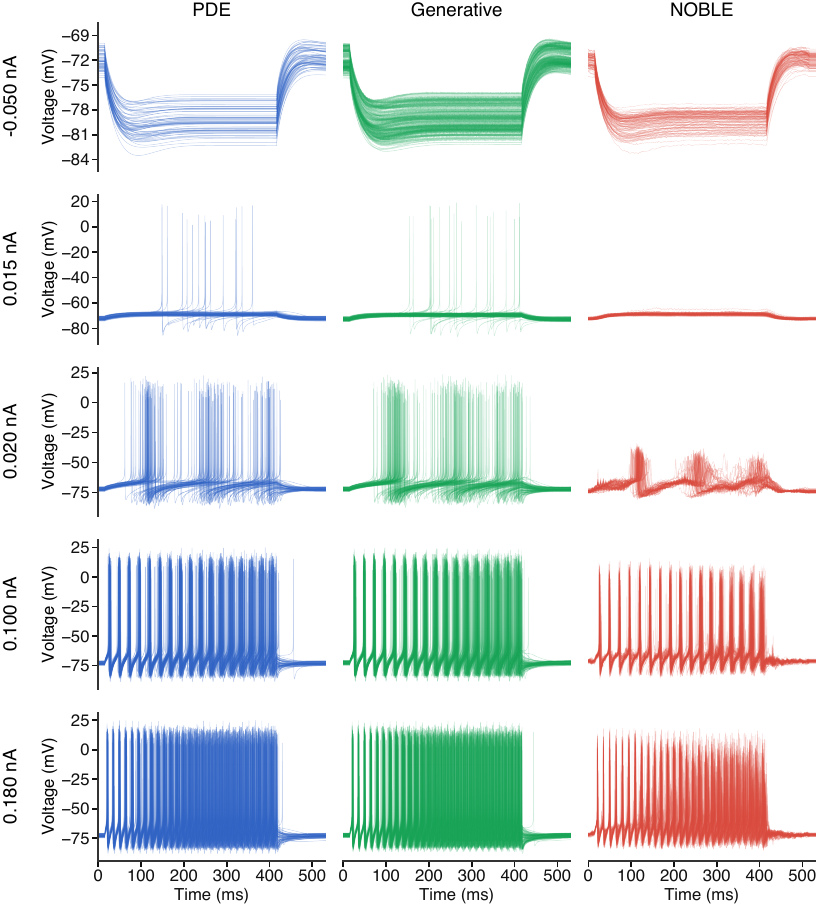}
    \caption{Examples of sets of reference and generated traces for square pulse input currents to the PVALB neuron at different amplitudes. From left to right, columns show voltage traces given by (1) PDE simulations, (2) our model, and (3) \texttt{NOBLE}.}
    \label{fig:pvalb_representative_traces}
\end{figure}

\begin{figure}[htbp]
    \subsection{Electrophysiological Feature Distribution Alignment}
    \vspace{0.2cm}
    \centering
    \includegraphics[width=\textwidth]{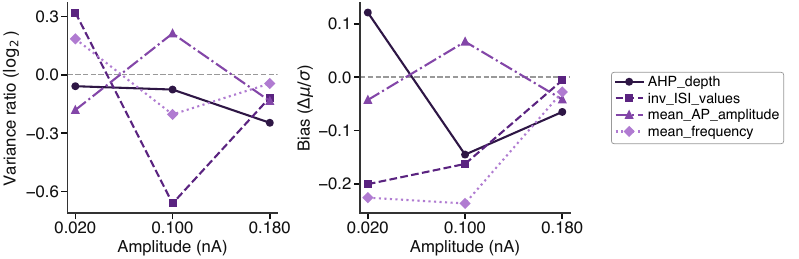}
    \caption{Variance ratio ($\log_2 \, \mathrm{Var(Generated) / Var(PDE)}$) and normalized mean bias $\Delta \mu / \sigma$ for PVALB feature distributions at selected amplitudes.}
    \label{fig:pvalb_distributional_alignment}
\end{figure}

\begin{figure}[htbp]
    \subsection{Frequency Sensitivity Comparison}
    \vspace{0.2cm}
    \centering
    \includegraphics[width=\textwidth]{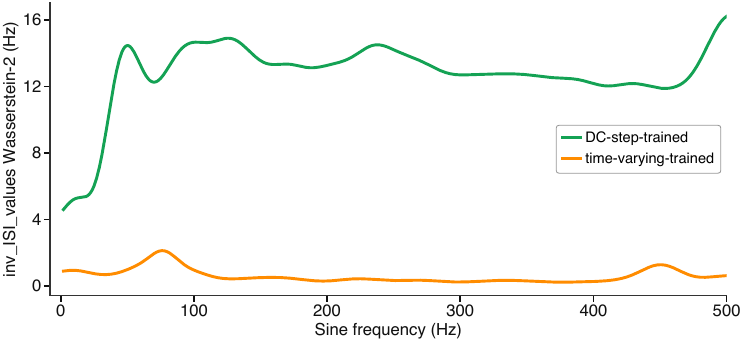}
    \caption{Wasserstein-2 distances between the reference and generated inv\_ISI\_values distributions for sine wave inputs of varying frequencies. Voltage responses are generated for sine waves with amplitudes 0.06nA, equilibrium position 0.05nA, and frequencies from 2 Hz to 500 Hz. Smoothed Wasserstein-2 distances are plotted for the DC-step-trained model (green) and the time-varying-trained model (orange).}
    \label{fig:pvalb_frequency_sweep}
\end{figure}

\begin{figure}[htbp]
    \subsection{Generalization to Complex Time-Varying Protocols}
    \label{app:more_rich_waveforms}
    \vspace{0.2cm}
    \centering
    \includegraphics[width=\textwidth,height=0.8\textheight,keepaspectratio]{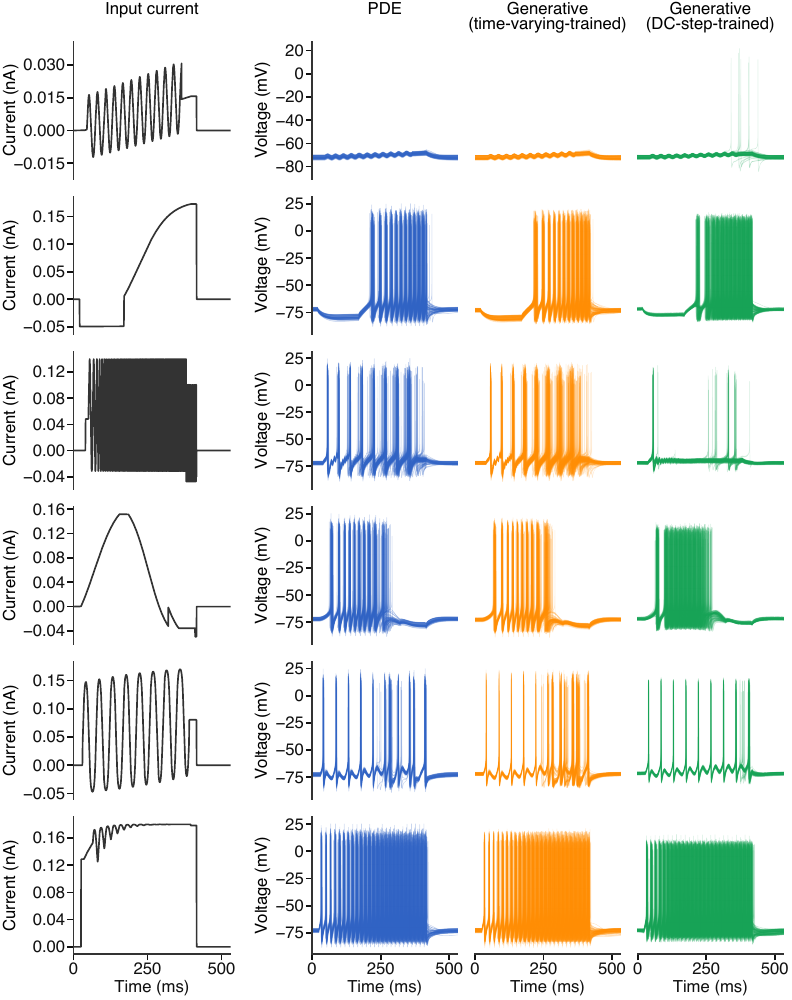}
    \caption{Additional examples of sets of reference and generated traces for unseen time-varying protocols. From left to right, columns show: (1) the input current trace, (2) the voltage traces generated by PDE simulations, (3) the voltage traces generated by our model when trained on time-varying current inputs, and (4) the voltage traces generated by our model when trained on DC step inputs.}
    \label{fig:pvalb_more_time_varying}
\end{figure}

\clearpage

\section{Supplementary Results for the VIP Neuron}\label{app:vip_results}

\begin{figure}[htbp]
    \subsection{Voltage Trace Ensembles}
    \vspace{0.2cm}
    \centering
    \includegraphics[width=\textwidth]{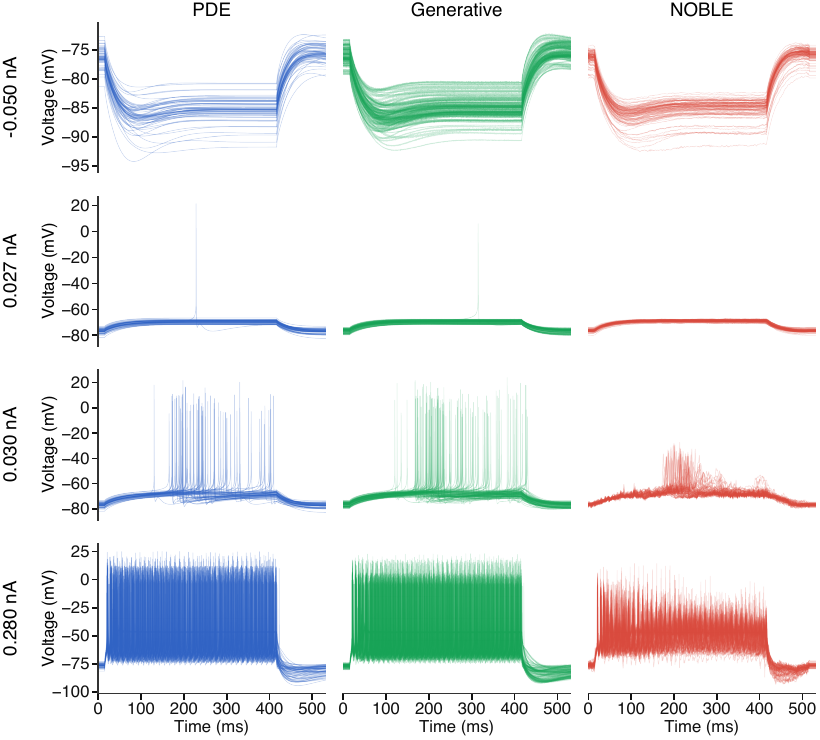}
    \caption{Examples of sets of reference and generated traces for square pulse input currents to the VIP neuron at different amplitudes. From left to right, columns show voltage traces given by (1) PDE simulations, (2) our model, and (3) \texttt{NOBLE}.}
    \label{fig:vip_representative_traces}
\end{figure}

\subsection{Performance Across Dynamical Regimes}
\begin{table}[H]
\centering
\begin{tabular}{lrrrrr}
\toprule
Regime
    & \hdr{AP\\width\\($\mu$s)}
    & \hdr{AHP\\depth\\(mV)}
    & \hdr{mean AP \\amplitude\\(mV)}
    & \hdr{inv ISI\\(Hz)}
    & \hdr{mean \\frequency\\(Hz)}
    \\
\midrule
All & 37.7 & 0.905 & 3.00 & 2.514 & 2.264 \\
Threshold & 39.2 & 1.903 & 5.91 & -- & 0.750 \\
Low-firing & 34.8 & 0.860 & 2.60 & 1.748 & 0.981 \\
High-firing & 39.3 & 0.907 & 3.16 & 2.939 & 3.009 \\
\bottomrule
\end{tabular}
\caption{Performance by dynamical regime, VIP neuron. Wasserstein-2 distances between reference and generated feature distributions, averaged across amplitudes in different regime types. Inv ISI is omitted in the threshold regime, where firing responses contain too few spikes to yield interspike intervals.}
\label{tab:vip_performance_by_dynamical_regime}
\end{table}

\subsection{Electrophysiological Feature Distributions}
\begin{figure}[H] 

    \vspace{0.2cm}
    \centering 
    \begin{minipage}[c]{0.65\textwidth} 
        \centering 
        \begin{overpic}[width=\linewidth]{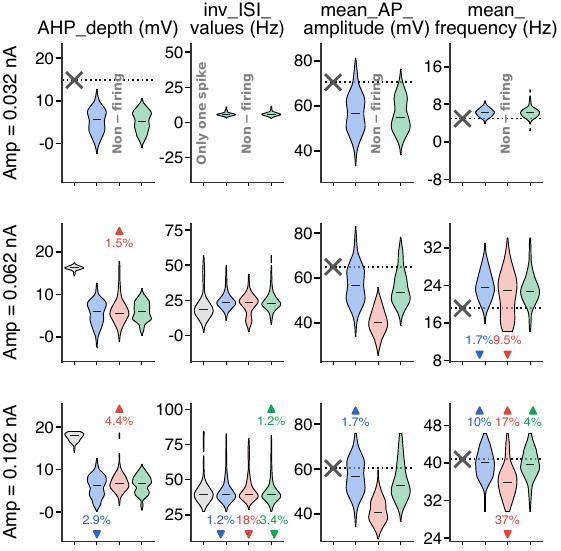}
            \put(0, 95){\textsf{\textbf{\LARGE A}}}
        \end{overpic}
    \end{minipage}% 
    \hfill
    \begin{minipage}[c]{0.33\textwidth} 
        \centering 
        \begin{overpic}[width=\linewidth]{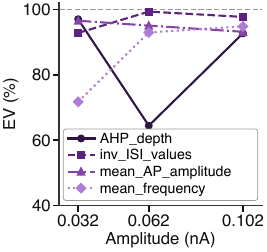}
            \put(-5, 89){\textsf{\textbf{\LARGE B}}}
        \end{overpic}
        
        \vspace{0.1cm} 
         
        \begin{overpic}[width=\linewidth]{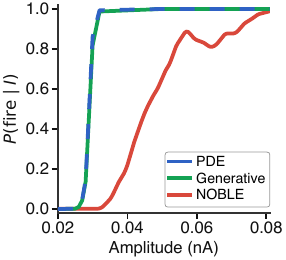}
            \put(-5, 89){\textsf{\textbf{\LARGE C}}}
        \end{overpic}
    \end{minipage} 
     
    \caption{\textbf{A)} Distribution of key electrophysiological features for a VIP neuron from experimental recordings (grey), PDE simulations (blue), \texttt{NOBLE} (red), and our model (green) across three representative amplitudes. Arrows indicate the fraction of values outside of the displayed axis range. \textbf{B)} Explained variance (EV, \%) across quantile levels of generated and PDE distributions. \textbf{C)} $P(\mathrm{fire} \mid I)$, the empirical firing probability given current amplitude for PDE simulations (blue; dashed), \texttt{NOBLE} (red), and our model (green).} 
    \label{fig:vip_combined} 
\end{figure}

\subsection{Electrophysiological Feature Distributions Alignment}

\begin{figure}[H]
    \vspace{0.2cm}
    \centering
    \includegraphics[width=\textwidth]{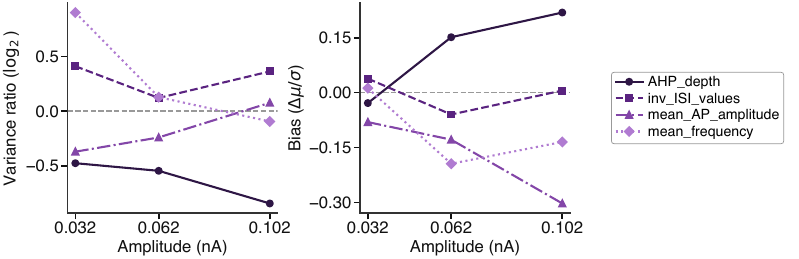}
    \caption{Variance ratio ($\log_2 \, \mathrm{Var(Generated) / Var(PDE)}$) and normalized mean bias $\Delta \mu / \sigma$ for VIP feature distributions at selected amplitudes.}
    \label{fig:vip_distributional_alignment}
\end{figure}

\end{document}